%% file: root.tex
\documentclass[letterpaper, 10 pt, conference]{ieeeconf}

\IEEEoverridecommandlockouts 
\usepackage[USenglish]{babel}
\usepackage{csquotes}
\usepackage[usenames,svgnames,table,rgb,dvipsnames]{xcolor} 
\usepackage{graphicx}
\graphicspath{{./graphics/}} 
\usepackage{tikz}
\usepackage{pgf}
\usepackage{subcaption} 
\usepackage{wrapfig}
\usepackage[]{mathtools} 
\usepackage{tabularx}
\usepackage{booktabs}
\usepackage{colortbl} 
\usepackage{makecell} 
\usepackage{duckuments}
\usepackage[lined,linesnumbered,ruled,noend]{algorithm2e}
\usepackage{placeins} 

\usepackage[nolist, nohyperlinks]{acronym}
\input{fMRT_acronyms}
\input{fMRT_macros}

\usepackage{url} 

\title{\LARGE \bf
Distilled Domain Randomization
}

\author{Julien~Brosseit$^{*,1}$ and Benedikt~Hahner$^{*,1}$ and Fabio~Muratore$^{*,1,2}$\\ and Michael~Gienger$^{2}$ and Jan~Peters$^{1}$
\thanks{* equal contribution}%
\thanks{$^{1}$Julien~Brosseit, Benedikt~Hahner, Fabio~Muratore, and Jan~Peters are with the Intelligent Autonomous Systems Group, Technical University of Darmstadt, Germany. \{julien.brosseit,benedikt.hahner\}@stud.tu-darmstadt.de, fabio@robot-learning.de}%
\thanks{$^{2}$Fabio~Muratore and Michael~Gienger are with the Honda Research Institute Europe, Offenbach am Main, Germany.}%
}%

\begin{document}

\maketitle
\thispagestyle{empty} 
\pagestyle{empty} 

\input{content/abstract}

\section{Introduction}
\label{sec:introduction}
\input{content/introduction}

\section{Related Work}
\label{sec:related_work}
\input{content/related_work}

\section{Problem Statement and Notation}
\label{sec:perliminaries}
\input{content/preliminaries}

\section{\acf{DiDoR}}
\label{sec:DiDoR}
\input{content/method}

\section{Experiments}
\label{sec:experiments}
\input{content/experiments}

\section{Conclusion}
\label{sec:conclusion}
\input{content/conclusion}



\iftrue
\section*{Appendix}
\input{content/appendix}
\fi

\section*{Acknowledgment}
Fabio Muratore gratefully acknowledges the financial support from \acl{HRIE}.
Jan Peters received funding from the European Union’s Horizon 2020 research and innovation programme under grant agreement No 640554.
The authors thank Michael Lutter for providing the picture of the cart-pole.
Calculations for this research were conducted on the Lichtenberg high performance computer of the TU Darmstadt.

\FloatBarrier

\bibliographystyle{./IEEEtran} 
\bibliography{./IEEEabrv,./IEEEexample,./references}

\end{document}

%% file: fMRT_acronyms.tex
\acrodef{AALR}{Amortized Approximate Likelihood Ratio}
\acrodef{ABC}{Approximate Bayesian Computation}
\acrodef{ADaPT}{Adaptive Policy Transfer for Stochastic Dynamics}
\acrodef{ADN}{Activation Dynamics Network}
\acrodef{ADR}{Active Domain Randomization}
\acrodef{ARC-t}{Asymmetric Regularized Cross-domain transformation}
\acrodef{A2RP}{Averaged Two-Replication Procedure}
\acrodef{APT}{Automatic Posterior Transformation}
\acrodef{ARPL}{Adversarially Robust Policy Learning}
\acrodef{BO}{Bayesian Optimization}
\acrodef{BayRn}{Bayesian Domain Randomization}
\acrodef{CAD}{Computer Aided Design}
\acrodef{CIO}{Contact-Invariant Optimization}
\acrodef{CMA}{Covariance Matrix Adaptation}
\acrodef{CMA-ES}{Covariance Matrix Adaptation - Evolutionary Strategies}
\acrodef{CMDP}{Contextual Markov Decision Process}
\acrodefplural{CMPD}[CMPDs]{Contextual Markov Decision Processes}
\acrodef{CNN}{Convolutional Neural Network}
\acrodef{CoM}{Center of Mass}
\acrodef{CVaR}{Conditional Value at Risk}
\acrodef{DA}{Domain Adaptation}
\acrodef{DAN}{Deep Adaptation Network}
\acrodef{DDPG}{Deep Deterministic Policy Gradient}
\acrodef{DiDoR}{Distilled Domain Randomization}
\acrodef{DISCO}{Double likelihood-free Inference Stochastic COntrol}
\acrodef{DMP}{Dynamic Movement Primitive}
\acrodefplural{DMP}[DMPs]{Dynamic Movement Primitives}
\acrodef{DNN}{Deep Neural Network}
\acrodef{DoF}{Degree of Freedom}
\acrodefplural{DOF}[DOF]{Degrees of Freedom}
\acrodef{DR}{Domain Randomization}
\acrodef{DRL}{Deep Reinforcement Learning}
\acrodef{DTW}{Dynamic Time Warping}
\acrodef{DS}{Dynamical System}
\acrodef{EoM}{Equation of Motion}
\acrodefplural{EoM}[EoMs]{Equations of Motion}
\acrodef{EEA}{Estimation Exploration Algorithm}
\acrodef{EPOpt}{Ensemble Policy Optimization}
\acrodef{FNN}{Feedforward Neural Network}
\acrodef{FPO}{Fingerprint Policy Optimization}
\acrodef{GA}{Genetic Algorithm}
\acrodef{GAN}{Generative Adversarial Network}
\acrodef{GMM}{Gaussian Mixture Model}
\acrodef{GP}{Gaussian Process}
\acrodefplural{GP}[GPs]{Gaussian Processes}
\acrodef{GPS}{Guided Policy Search}
\acrodef{GPU}{Graphics Processing Unit}
\acrodef{HER}{Hindsight Experience Replay}
\acrodef{HRI}{Honda Research Institute}
\acrodef{HRIE}{Honda Research Institute Europe}
\acrodef{I2RP}{Independent Two-Replication Procedure}
\acrodef{IAS}{Intelligent Autonomous Systems}
\acrodef{IEEE}{Institute of Electrical and Electronics Engineers}
\acrodef{IID}{Independent and Identically Distributed}
\acrodef{KL}{Kullback-Leibler}
\acrodef{LFI}{Likelihood-Free Inference}
\acrodef{LTI}{Linear Time-Invariant}
\acrodef{LQR}{Linear-Quadratic Regulator}
\acrodef{LSDA}{Large Scale Detection through Adaptation}
\acrodef{LSTM}{Long Short-Term Memory}
\acrodef{LWPR}{Locally Weighted Projection Regression}
\acrodef{MAF}{Masked Autoregressive Flow}
\acrodef{MCMC}{Markov Chain Monte Carlo}
\acrodef{MDN}{Mixture Density Network}
\acrodef{MDP}{Markov Decision Process}
\acrodefplural{MDP}{Markov Decision Processes}
\acrodef{MoG}{Mixture of Gaussians}
\acrodef{MP}{Movement Primitive}
\acrodef{MPC}{Model Predictive Control}
\acrodef{MPPI}{Model Predictive Path Integral}
\acrodef{MSE}{Mean-Squared Error}
\acrodef{NN}{Neural Network}
\acrodef{NPDR}{Neural Posterior Domain Randomization}
\acrodef{NPG}{Natural Policy Gradient}
\acrodef{ODE}{Ordinary Differential Equation}
\acrodef{ODEphys}[ODE]{Open Dynamics Engine}
\acrodef{OG}{Optimality Gap}
\acrodef{PNN}{Progressive Neural Network}
\acrodef{PD}{Proportional-Derivative}
\acrodef{PDDR}{Policy Distillation with Domain Randomization}
\acrodef{POMDP}{Partially-Observable Markov Decision Process}
\acrodef{PoWER}{Policy learning by Weighting Exploration with the Returns}
\acrodef{PPO}{Proximal Policy Optimization}
\acrodef{P2PDRL}{Peer-to-Peer Distillation Reinforcement Learning}
\acrodef{ProMP}{Probabilistic Movement Primitive}
\acrodef{QBB}{Quanser Ball-Balancer}
\acrodef{QCP}{Quanser Cart-Pole}
\acrodef{QQ}{Quanser Qube}
\acrodef{RA}{Retrospective Approximation}
\acrodef{RARL}{Robust Adversarial Reinforcement Learning}
\acrodef{RBF}{Radial Basis Function}
\acrodef{REPS}{Relative Entropy Policy Search}
\acrodef{ResNet}{Residual neural network}
\acrodef{RFF}{Random Fourier Features}
\acrodef{RGB}{Red-Green-Blue}
\acrodef{RGBD}{Red-Green-Blue-Depth}
\acrodef{RMSE}{Root-Mean-Square Error}
\acrodef{RNN}{Recurrent Neural Network}
\acrodef{RL}{Reinforcement Learning}
\acrodef{SAA}{Sample Average Approximation}
\acrodef{SMC}{Sequential Monte Carlo}
\acrodef{SNLE}{Sequential Neural Likelihood Estimation}
\acrodef{SNPE}{Sequential Neural Posterior Estimation}
\acrodef{SNRE}{Sequential Neural Ratio Estimation}
\acrodef{SOB}{Simulation Optimization Bias}
\acrodef{SP}{Stochastic Program}
\acrodefplural{SP}{Stochastic Programs}
\acrodef{SPOTA}{Simulation-based Policy Optimization with Transferability Assessment}
\acrodef{SPRL}{Self-Paced Reinforcement Learning}
\acrodef{SRCC}{Sim-vs-Real Correlation Coefficient}
\acrodef{SVPG}{Stein Variational Policy Gradient}
\acrodef{SysId}{System Identification}
\acrodef{TRPO}{Trust Region Policy Optimization}
\acrodef{UCB}{Upper Confidence Bound}
\acrodef{UDR}{Uniform Domain Randomization}
\acrodef{UCBOG}{Upper Confidence Bound on the Optimality Gap}
\acrodef{UCSOB}{Upper Confidence bound on the Simulation Optimization Bias}
\acrodef{UP-OSI}{Universal Policy - Online System Identification}
\acrodef{VAE}{Variational AutoEncoder}

%% file: fMRT_macros.tex
\makeatletter
\@ifpackageloaded{xparse}{}{\usepackage{xparse}}
\@ifpackageloaded{tabularx}{}{\usepackage{tabularx}} 
\@ifpackageloaded{xifthen}{}{\usepackage{xifthen}}
\@ifpackageloaded{xspace}{}{\usepackage{xspace}} 
\@ifpackageloaded{amsopn}{}{\usepackage{amsopn}} 
\@ifpackageloaded{amssymb}{}{\usepackage{amssymb}} 
\@ifpackageloaded{mathtools}{}{\usepackage{mathtools}} 
\@ifpackageloaded{bm}{}{\usepackage{bm}} 
\@ifpackageloaded{siunitx}{}{%
\usepackage[detect-all,scientific-notation=false,separate-uncertainty=true]{siunitx} 
\sisetup{%
    retain-explicit-plus=true,
    detect-weight=true,
    detect-family=true,
	output-exponent-marker=\ensuremath{\mathrm{e}},
	per-mode=symbol,
	range-phrase={,}\ ,
    range-units=brackets,
    open-bracket={[},
    close-bracket=],
}%
}
\makeatother

\definecolor{fMRTblack}{HTML}{2B2E34}
\definecolor{fMRTwhite}{HTML}{F3EEE1}
\definecolor{fMRTyellow}{HTML}{FFD564}
\definecolor{fMRTorange}{HTML}{FF5500}
\colorlet{fMRTlightGray}{white!80!fMRTblack}
\colorlet{fMRTgray}{white!35!fMRTblack}
\colorlet{fMRTdarkGray}{white!20!fMRTblack}
\colorlet{fMRTlightOrange}{fMRTorange!60!fMRTwhite}
\definecolor{fMRTblue}{HTML}{3333B3}
\colorlet{fMRTlightBlue}{fMRTblue!30!white}
\definecolor{fMRTverylightRed}{HTML}{FF8E6B}
\colorlet{fMRTlightRed}{red!60!fMRTwhite}
\colorlet{fMRTdarkRed}{red!70!fMRTgray} 
\definecolor{fMRTlightGreen}{HTML}{8CDD81} 
\definecolor{fMRTborwn}{HTML}{8B4513} 
\colorlet{fMRTlightBorwn}{fMRTborwn!60!fMRTwhite}

\colorlet{fMRTdark@backgroundInner}{fMRTblack}
\colorlet{fMRTdark@backgroundOuter}{fMRTblack}
\colorlet{fMRTlight@backgroundInner}{fMRTwhite}
\colorlet{fMRTlight@backgroundOuter}{fMRTwhite}

\newcolumntype{L}{>{\raggedright\arraybackslash}X}
\newcolumntype{R}{>{\raggedleft\arraybackslash}X}
\newcolumntype{C}{>{\centering\arraybackslash}X}
\newcolumntype{P}[1]{>{\raggedright\arraybackslash}p{#1}} 

\newcommand{\optsym}{\star} 
\newcommand{\opt}{^\optsym}
\newcommand{\obssym}{\text{obs}}
\newcommand{\initsym}{\text{init}}

\newcommand{\simusym}{\text{sim}}

\newcommand{\realsym}{\text{real}}
\newcommand{\real}{^\realsym}

\newcommand{\RR}{\mathbb{R}} 

\DeclareDocumentCommand{\set}{O{} O{} m}{\{#3\}_{#1}^{#2}} 
\DeclareDocumentCommand{\tuple}{O{} O{} m}{\left\langle#3\right\rangle_{#1}^{#2}} 

\newcommand{\stateset}[1][]{\mathcal{S}_{#1}} 
\newcommand{\actionset}[1][]{\mathcal{A}_{#1}} 
\newcommand{\transprobsym}{\mathcal{P}}
\makeatletter
\DeclareDocumentCommand{\@transprob}{O{} O{} m}{\transprobsym_{#1}^{#2}\left( #3\right) }  
\DeclareDocumentCommand{\@@transprob}{O{} O{} m}{\transprobsym_{#1}^{#2}\!\left( #3\right) }
\newcommand{\transprob}{\@ifstar\@transprob\@@transprob}
\makeatother
\newcommand{\initstatedistrsym}{\mu}
\makeatletter
\DeclareDocumentCommand{\@initstatedistr}{O{} O{} m}{\initstatedistrsym_{#1}^{#2}\left( #3\right) }  
\DeclareDocumentCommand{\@@initstatedistr}{O{} O{} m}{\initstatedistrsym_{#1}^{#2}\!\left( #3\right) }
\newcommand{\initstatedistr}{\@ifstar\@initstatedistr\@@initstatedistr}
\makeatother
\newcommand{\mdp}[1][]{\mathcal{M}_{#1}} 
\newcommand{\dimstate}{{n_s}} 
\newcommand{\dimact}{{n_a}} 
\newcommand{\dimpolparam}{{n_{\polparam}}} 
\newcommand{\dimdomparam}{{n_{\domparam}}} 

\newcommand{\candsym}{{n_c}}
\newcommand{\refsym}{{n_r}}

\newcommand{\eg}{e.g.\xspace}

\newcommand{\wrt}{w.r.t.~}
\newcommand{\aka}{a.k.a.~}

\newcommand{\Simtosim}{{Sim-to-sim}\xspace}
\newcommand{\simtosim}{{sim-to-sim}\xspace}

\newcommand{\Simtoreal}{{Sim-to-real}\xspace}
\newcommand{\simtoreal}{{sim-to-real}\xspace}

\newcommand{\distr}[3]{#1\left( #2 \big| #3\right) } 
\newcommand{\distrnormal}[2]{\distr{\mathcal{N}}{#1}{#2}} 
\newcommand{\distruniform}[2]{\distr{\mathcal{U}}{#1}{#2}} 
\newcommand{\Esub}[2]{\mathbb{E}_{#1} \! \left[ #2\right] } 

\makeatletter
\newcommand{\@KL}[2]{\mathrm{KL}(#1 \Vert #2)} 
\newcommand{\@@KL}[3][]{\mathrm{KL}\!\left( #2 #1\Vert #3\right)} 
\newcommand{\KL}{\@ifstar\@KL\@@KL}
\makeatother
\newcommand{\KLlosssym}{\mathcal{L}_{\textrm{KL}}}
\newcommand{\quantilesym}{Q} 

\renewcommand{\exp}[1]{\mathrm{exp}\left( #1\right) } 

\makeatletter
\newcommand{\@given}[2]{\left. #1\right| #2} 
\newcommand{\@@given}[3][]{#2 #1| #3} 
\newcommand{\given}{\@ifstar\@given\@@given}
\makeatother

\DeclareDocumentCommand{\distmeas}{O{} O{} m}{\text{D}_{#1}^{#2} \!\left( #3\right)} 

\newcommand{\rewsym}{r}
\DeclareDocumentCommand{\rewfcn}{O{} O{} m}{\rewsym_{#1}^{#2}\!\left( #3\right) }

\newcommand{\trajssym}{\bm{\tau}}
\newcommand{\trajspace}{\mathcal{T}}

\DeclareDocumentCommand{\traj}{O{} O{}}{\trajssym_{#1}^{#2}} 
\newcommand{\statedistrsym}{\mu}
\DeclareDocumentCommand{\statedistr}{O{} O{} m}{\statedistrsym_{#1}^{#2}\!\left( #3\right) } 

\newcommand{\domparam}{\xi}
\newcommand{\domparams}{\bm{\xi}}

\newcommand{\domparamdistrsym}{p} 
\makeatletter
\DeclareDocumentCommand{\@domparamdistr}{O{} O{} m}{\domparamdistrsym_{#1}^{#2}\left( #3\right) }  
\DeclareDocumentCommand{\@@domparamdistr}{O{} O{} m}{\domparamdistrsym_{#1}^{#2}\!\left( #3\right) }
\newcommand{\domparamdistr}{\@ifstar\@domparamdistr\@@domparamdistr}
\makeatother

\newcommand{\domdistrparams}{\bm{\phi}} 

\newcommand{\densestsym}{q}
\DeclareDocumentCommand{\densest}{O{} O{} m}{\densestsym_{#1}^{#2}\!\left( #3\right)} 

\DeclareDocumentCommand{\domparamprior}{O{} O{} m}{\domparamdistrsym_{#1}^{#2}\!\left( #3\right) } 
\DeclareDocumentCommand{\domparamproposal}{O{} O{} m}{\tilde{\domparamdistrsym}_{#1}^{#2}\!\left( #3\right) } 
\DeclareDocumentCommand{\domparamlikelihood}{O{} O{} m}{\domparamdistrsym_{#1}^{#2}\!\left( #3\right) } 
\DeclareDocumentCommand{\domparamposterior}{O{} O{} m}{\domparamdistrsym_{#1}^{#2}\!\left( #3\right) } 
\DeclareDocumentCommand{\estdomparamposterior}{O{} O{} m}{\hat{\domparamdistrsym}_{#1}^{#2}\!\left( #3\right) } 

\newcommand{\genmodelsym}{g}
\DeclareDocumentCommand{\genmodel}{O{} O{} m}{\genmodelsym{#1}^{#2}\!\left( #3\right) }

\newcommand{\summstatsym}{f}

\DeclareDocumentCommand{\summstatfcn}{O{} O{} m}{\summstatsym_{#1}^{#2}\!\left( #3\right) }

\newcommand{\polsym}{\pi}
\DeclareDocumentCommand{\pol}{O{} O{} m}{\polsym_{#1}^{#2}\!\left( #3\right) } 
\newcommand{\polparam}{\theta}

\DeclareDocumentCommand{\polparams}{O{} O{}}{\ftheta_{#1}^{#2}} 
\DeclareDocumentCommand{\polparamsopt}{O{} O{}}{\ftheta_{#1}^{#2 \optsym}} 
\DeclareDocumentCommand{\polparamopt}{O{} O{}}{\polparam_{#1}^{#2 \optsym}} 

\newcommand{\GPsym}{\mathcal{GP}}

\newcommand{\acqfcnsym}{a}
\DeclareDocumentCommand{\acqfcn}{O{} O{} m}{\acqfcnsym_{#1}^{#2}\!\left( #3\right) }

\newcommand{\optgapsym}{G}
\DeclareDocumentCommand{\optgap}{O{} O{} m}{\optgapsym_{#1}^{#2}\!\left( #3\right) } 
\newcommand{\estoptgapsym}{\hat{G}}
\DeclareDocumentCommand{\estoptgap}{O{} O{} m}{\estoptgapsym_{#1}^{#2}\!\left( #3\right) } 
\newcommand{\meanoptgapsym}{\bar{G}}
\DeclareDocumentCommand{\optgapmean}{O{} O{} m}{\meanoptgapsym_{#1}^{#2}\!\left( #3\right) } 
\newcommand{\optgapsampsetsym}{\mathcal{G}}
\DeclareDocumentCommand{\optgapsampset}{O{} O{}}{\optgapsampsetsym_{#1}^{#2}} 

\DeclareDocumentCommand{\studpolparams}{O{} O{}}{ \polparams[#1][s #2] }
\DeclareDocumentCommand{\teachpolparams}{O{} O{}}{ \polparams[#1][t #2] }

\iftrue
\newcommand{\bootsym}{B}
\DeclareDocumentCommand{\bootestoptgap}{O{} O{} m}{\estoptgapsym_{#1}^{\bootsym #2}\!\left( #3\right) }
\DeclareDocumentCommand{\bootmeanoptgap}{O{} O{} m}{\meanoptgapsym_{#1}^{\bootsym #2}\!\left( #3\right) }
\DeclareDocumentCommand{\bootoptgapsampset}{O{} O{}}{\optgapsampsetsym_{#1}^{\bootsym #2}}
\fi

\newcommand{\edrsym}{J}
\DeclareDocumentCommand{\edr}{O{} O{} m}{\edrsym_{#1}^{#2}\!\left( #3\right)} 

\newcommand{\estedrsym}{\hat{J}}
\DeclareDocumentCommand{\estedr}{O{}  O{} m}{\estedrsym_{#1}^{#2}\!\left( #3\right)} 
\newcommand{\eedrsym}{J} 
\DeclareDocumentCommand{\eedr}{O{} O{} m}{\eedrsym_{#1}^{#2}\!\left( #3\right)} 
\DeclareDocumentCommand{\esteedr}{O{} O{} m}{\estedrsym_{#1}^{#2}\!\left( #3\right)} 
\newcommand{\objfunsym}{f}
\DeclareDocumentCommand{\objfun}{O{} m}{\objfunsym_{#1}\!\left( #2\right)} 

\newcommand{\dsdynsym}{f}
\DeclareDocumentCommand{\dsdyn}{O{} m}{\dsdynsym_{#1}\!\left( #2\right)} 

\newcommand{\potdynsym}{f}
\newcommand{\potsdynsym}{\ff}
\DeclareDocumentCommand{\potdyn}{O{} m}{\potdynsym_{#1}\!\left( #2\right)} 
\DeclareDocumentCommand{\potsdyn}{O{} m}{\potsdynsym_{#1}\!\left( #2\right)} 

\newcommand{\taskpos}{x}

\newcommand{\taskposs}{\bm{x}}

\newcommand{\tasksposs}{\bm{X}}

\ExplSyntaxOn
\NewDocumentCommand{\eqsref}{m}{\quinn_eqsref:n {#1}}
\seq_new:N \l_quinn_eqsref_seq
\cs_new:Npn \quinn_eqsref:n #1
{
	\seq_set_split:Nnn \l_quinn_eqsref_seq { , } { #1 }
	\seq_pop_right:NN \l_quinn_eqsref_seq \l_tmpa_tl
	( 
	\seq_map_inline:Nn \l_quinn_eqsref_seq
	{ \ref{##1},\nobreakspace } 
	\exp_args:NV \ref \l_tmpa_tl 
	) 
}
\ExplSyntaxOff

\SetKw{kwNot}{not}
\SetKw{kwAnd}{and}
\SetKw{kwBreak}{break}
\SetKwInOut{Input}{input}
\SetKwInOut{Output}{output}
\SetKwInOut{Initialization}{init}
\SetKwComment{Comment}{$\triangleright$\ }{}

\SetCommentSty{famuracommfont}
\SetKwRepeat{Do}{do}{while} 
\SetKwBlock{With}{with}{end} 
\newcommand{\llIf}[2]{{\let\par\relax\lIf{#1}{#2}}}
\newcommand{\llElse}[1]{{\let\par\relax\lElse{#1}}}
\newcommand{\llFor}[2]{{\let\par\relax\lFor{#1}{#2}}}

\newcommand{\PolOptsym}{\textsc{PolOpt}}
\newcommand{\DistrOptsym}{\textsc{DistrOpt}}
\newcommand{\Infersym}{\textsc{Infer}}

\DeclareBoldMathCommand{\fnull}{0}
\DeclareBoldMathCommand{\fO}{0}
\DeclareBoldMathCommand{\fA}{A}
\DeclareBoldMathCommand{\fa}{a}
\DeclareBoldMathCommand{\fB}{B}
\DeclareBoldMathCommand{\fb}{b}
\DeclareBoldMathCommand{\fC}{C}
\DeclareBoldMathCommand{\fc}{c}
\DeclareBoldMathCommand{\fD}{D}
\DeclareBoldMathCommand{\fd}{d}
\DeclareBoldMathCommand{\fE}{E}
\DeclareBoldMathCommand{\fe}{e}
\DeclareBoldMathCommand{\fF}{F}
\DeclareBoldMathCommand{\ff}{f}
\DeclareBoldMathCommand{\fG}{G}
\DeclareBoldMathCommand{\fg}{g}
\DeclareBoldMathCommand{\fH}{H}
\DeclareBoldMathCommand{\fh}{h}
\DeclareBoldMathCommand{\fI}{I}
\DeclareBoldMathCommand{\fJ}{J}
\DeclareBoldMathCommand{\fK}{K}
\DeclareBoldMathCommand{\fM}{M}
\DeclareBoldMathCommand{\fm}{m}
\DeclareBoldMathCommand{\fN}{N}
\DeclareBoldMathCommand{\fn}{n}
\DeclareBoldMathCommand{\fo}{o}
\DeclareBoldMathCommand{\fP}{P}
\DeclareBoldMathCommand{\fp}{p}
\DeclareBoldMathCommand{\fQ}{Q}
\DeclareBoldMathCommand{\fq}{q}
\DeclareBoldMathCommand{\fq}{q}

\DeclareBoldMathCommand{\fR}{R}
\DeclareBoldMathCommand{\fr}{r}

\DeclareBoldMathCommand{\fS}{S}
\DeclareBoldMathCommand{\fs}{s}

\DeclareBoldMathCommand{\ft}{t}
\DeclareBoldMathCommand{\fT}{T}
\DeclareBoldMathCommand{\fU}{U}
\DeclareBoldMathCommand{\fu}{u}
\DeclareBoldMathCommand{\fV}{V}
\DeclareBoldMathCommand{\fv}{v}
\DeclareBoldMathCommand{\fW}{W}
\DeclareBoldMathCommand{\fw}{w}
\DeclareBoldMathCommand{\fx}{x}
\DeclareBoldMathCommand{\fz}{z}

\newcommand{\fxbar}{\bar{\fx}}

\DeclareBoldMathCommand{\fY}{Y}
\DeclareBoldMathCommand{\fy}{y}
\DeclareBoldMathCommand{\fZ}{Z}
\DeclareBoldMathCommand{\falpha}{\alpha}

\DeclareBoldMathCommand{\fbeta}{\beta}
\DeclareBoldMathCommand{\fchi}{\chi}
\DeclareBoldMathCommand{\fepsilon}{\epsilon}
\DeclareBoldMathCommand{\fvarepsilon}{\varepsilon}
\DeclareBoldMathCommand{\fgamma}{\gamma}
\DeclareBoldMathCommand{\fGamma}{\Gamma}
\DeclareBoldMathCommand{\flambda}{\lambda}
\DeclareBoldMathCommand{\fLambda}{\Lambda}
\DeclareBoldMathCommand{\fmu}{\mu}
\DeclareBoldMathCommand{\fnu}{\nu}
\DeclareBoldMathCommand{\fomega}{\omega}
\DeclareBoldMathCommand{\fpi}{\pi}
\DeclareBoldMathCommand{\fphi}{\phi}
\DeclareBoldMathCommand{\fPhi}{\Phi}
\DeclareBoldMathCommand{\fpsi}{\psi}
\DeclareBoldMathCommand{\fsigma}{\sigma}
\DeclareBoldMathCommand{\fSigma}{\Sigma}
\DeclareBoldMathCommand{\ftau}{\tau}
\DeclareBoldMathCommand{\ftheta}{\theta}
\DeclareBoldMathCommand{\fTheta}{\Theta}

\DeclareBoldMathCommand{\fxi}{\xi}

%% file: content/abstract.tex
\begin{abstract}
Deep reinforcement learning is an effective tool to learn robot control policies from scratch.
However, these methods are notorious for the enormous amount of required training data which is prohibitively expensive to collect on real robots.
A highly popular alternative is to learn from simulations, allowing to generate the data much faster, safer, and cheaper.
Since all simulators are mere models of reality, there are inevitable differences between the simulated and the real data, often referenced as the \enquote*{reality gap}.
To bridge this gap, many approaches learn one policy from a distribution over simulators.
In this paper, we propose to combine reinforcement learning from randomized physics simulations with policy distillation.
Our algorithm, called \ac{DiDoR}, distills so-called teacher policies, which are experts on domains that have been sampled initially, into a student policy that is later deployed.
This way, \ac{DiDoR} learns controllers which transfer directly from simulation to reality, i.e., without requiring data from the target domain.
We compare \ac{DiDoR} against three baselines in three sim-to-sim as well as two sim-to-real experiments.
Our results show that the target domain performance of policies trained with \ac{DiDoR} is en par or better than the baselines'.
Moreover, our approach neither increases the required memory capacity nor the time to compute an action, which may well be a point of failure for successfully deploying the learned controller.
\end{abstract}

%% file: content/introduction.tex
Learning from randomized simulations has shown to be a promising approach for learning robot control policies that transfer to the real world. Examples cover manipulation \cite{OpenAI_18,Peng_etal_2018,Lowrey_etal_2018,OpenAI_19,Chebotar_Fox_19}, trajectory optimization~\cite{Mordatch_Todorov_15}, continuous control~\cite{Yu_Turk_17,Muratore_Peters_21_PAMI,Muratore_Peters_21_RAL}, vision~\cite{Tobin_Abbeel_17,Sadeghi_Levine_17,Tobin_Abbeel_18,James_Bousmalis_19}, and locomotion tasks~\cite{Tan_Vanhoucke_18,Antonova_Rai_Kragic_19,Peng_Levine_20}.
Independent of the task, all domain randomization methods can be classified based on the fact if they use target domain data to update the distribution over simulators or not.
While adapting the domain parameter distribution is in general superior, there are problem settings where (i) the target domain is not accessible until deployment, (ii)  iterating a \simtoreal loop is too expensive, or (iii) the problem can be simulated well enough to directly execute the learned policy.
For any of these three scenarios, zero-shot learning using domain randomization is either a viable or the only option.
However, naive randomization approaches have the disadvantage that the domain parameter distribution is easily selected too wide, such that the agent can not learn something meaningful.
In this case, the cause of failure is the large variance of the policy update, which results from collecting training data in many different environments at every update step. 
To resolve this issue, we propose to learn individual expert policies each mastering exactly one domain which has been sampled randomly from the distribution over simulators.
Subsequently, these experts are used as teachers from which a student can learn by imitation.
Here, we focus on the supervised learning technique called on-policy distillation~\cite{Ross_Bagnell_11,parisotto_2015,Lin_Zhou_17,Czarnecki_etal_2019} which belongs to the overarching category of knowledge distillation~\cite{hinton_distilling_2015} methods.
Policy distillation transfers the behavior of one or more teacher policies, typically deep neural networks, into one (smaller) student policy.
There are several other ways to learn from an ensemble of teachers~\cite{Osa_Peters_18}, which are out of the scope of this paper.

\begin{figure}[t]
    \centering

    \includegraphics[width=\columnwidth]{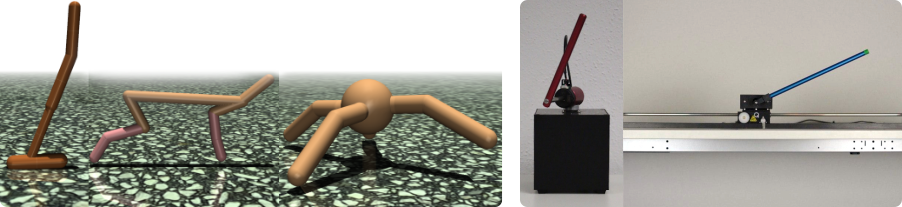}
    \caption{%
    The evaluation platforms.
    \Simtosim (left): hopper, half-cheetah, and ant.
    \Simtoreal (right): Furuta pendulum, and cart-pole.
    \vspace{-1em}
    }
    \label{fig:platforms}
\end{figure}

\paragraph*{Contributions}
We advance the state of the art by introducing \acf{DiDoR}, the first algorithm which leverages the synergies between domain randomization and policy distillation to bridge the reality gap in robotics.
Three particular synergies are:
(i) The individual training procedures are more stable and reproducible compared to other domain randomization approaches, because the reinforcement learning for the teachers is done for one simulation instance only.
(ii) For \ac{DiDoR}, the computer only needs to hold one policy in memory at a time.
(iii) Since the final network is small, the time to compute the next command is reduced significantly compared to methods that query an ensemble of experts.
Especially the last two points are beneficial to satisfy real-time requirements on a physical device.
Furthermore, \ac{DiDoR} is easy to implement and parallelize, yet effective as demonstrated by comparing against three baselines in three sim-to-sim and two sim-to-real experiments.
Finally, we made all our implementations available open-source in a common software framework\footnote{\url{https://github.com/MisterJBro/SimuRLacra}}.
This also includes \acl{P2PDRL}~\cite{Zhao2020}, which has not been tested in \simtoreal settings so far.

%% file: content/related_work.tex
We categorize the related work by the two techniques that make up \ac{DiDoR}: domain randomization (Section~\ref{sec:related_works_domain_randomization}), and policy distillation (Section~\ref{sec:related_works_policy_distillation}).

\subsection{Domain Randomization}
\label{sec:related_works_domain_randomization}
The core idea of domain randomization is to expose the agent to various instances of the same environment such that it learns a policy which solves the task for a distribution of domain parameters, and therefore is more robust against model mismatch.
Here, we focus on static domain randomization as used in \ac{DiDoR}.
These approaches do not update the domain parameter distribution, and directly deploy the learned policy in the target domain.
More specifically, we narrow our selection down to algorithms that randomize the system dynamics and achieved the \simtoreal transfer.

Most prominently, the robotic in-hand manipulation reported by OpenAI in \cite{OpenAI_18} demonstrated that domain randomization in combination with model engineering and the usage of recurrent neural networks enables direct sim-to-real transfer on an unprecedented difficulty level.
Similarly, Peng~et~al.~\cite{Peng_etal_2018} combined model-free \acs{RL} with recurrent neural network
policies trained using experience replay in order to push an object by controlling a robotic arm.
In contrast, Mordatch~et~al.~\cite{Mordatch_Todorov_15} employed finite model ensembles to run trajectory optimization on a small-scale humanoid robot.
After carefully identifying the system’s parameters, Lowrey~et~al.~\cite{Lowrey_etal_2018} learned a continuous controller for a three-finger positioning task.
Their results show that the policy learned from the identified model was able to perform the sim-to-real transfer, but the policies learned from an ensemble of models were more robust to modeling errors.
Tan~et~al.~\cite{Tan_Vanhoucke_18} presented an example for learning quadruped gaits from randomized simulations.
They found empirically that sampling domain parameters from a uniform distribution together with applying random forces and regularizing the observation space can be sufficient to cross the reality gap.
Aside from to the previous methods, Muratore~et~al.~\cite{Muratore_Peters_21_PAMI} introduce an approach to estimate the transferability of a policy learned from randomized physics simulations.
Moreover, the authors propose a meta-algorithm which provides a probabilistic guarantee on the performance loss when transferring the policy between two domains form the same distribution.

\subsection{Policy Distillation}
\label{sec:related_works_policy_distillation}
Policy distillation describes the transfer of knowledge about a specific task from one or multiple teachers to a student.
This technique originates from the seminal work of Hinton~et~al.~\cite{hinton_distilling_2015} which introduced knowledge distillation.
The supervised learning method distills an ensemble of models into a single model, which typically is smaller than the original ones.
Therefore, knowledge distillation can be seen as a generalization of model compression.
In the context of (deep) \acs{RL}, the knowledge transfer is achieved by training the student to imitate the teacher such that the decision making of the policy matches.
While most of the related work considered deep Q-networks, knowledge distillation can be applied regardless of the model's architecture or the method used for training the teachers.
To the best of our knowledge, policy distillation was not applied in a \simtoreal scenario, yet.

Policy distillation is strongly connected to multi-task learning where each teacher is specialized in a different task, e.g. an Atari game~\cite{parisotto_2015,Rusu_etal_2016b}, or an image classifier~\cite{Li_Bilen_20}. 
In the multi-task  setting, the teachers' knowledge is distilled into a student in order to bootstrap the student's learning procedure.
Note that the approach presented in this paper has a fundamentally different goal.
\ac{DiDoR} distills multiple experts of the same task, but specialized in different instances, to obtain a control policy that is robust to model mismatch.
Romero~et~al.~\cite{Romero_2014} showed that besides model compression, policy distillation can yield students which improve over their teachers' performance by choosing deeper and thinner neural networks for the student together with reusing the teachers' hidden layers.
%
%
Recently, Zhao~et~al.~\cite{Zhao2020} proposed an online peer-to-peer distillation called \acf{P2PDRL} which differs from previous methods during the distillation step.
In line with the idea of domain randomization, each teacher is assigned to a random domain and the policy update is regularized by its peers.
The regularization reduces the variance of the policy update.
Unlike \ac{DiDoR}, the teachers are assigned to a new domain at every iteration of the algorithm, which makes the training of the domain experts more noisy for \ac{P2PDRL}.
Moreover, \ac{P2PDRL} requires all teachers to be held in memory during training.
The authors evaluated \ac{P2PDRL} on five \simtosim experiments using the MuJoCo physics engine.
Czarnecki~et~al.~\cite{Czarnecki_etal_2019} provided a unifying overview of policy distillation.
According to their categorization, \ac{DiDoR} can be labeled an \emph{on-policy distillation} method.
The authors emphasize that seemingly minor differences between the approaches can have major effects on the resulting algorithm, such as the loss of its convergence guarantees, or a different application scenario.

%% file: content/preliminaries.tex
Consider a discrete-time dynamical system
\begin{equation}
\begin{gathered}
	\label{eq_sbi_mdp_def}
	\fs_{t+1} \sim \transprob[\domparams]{\given{\fs_{t+1}}{\fs_t, \fa_t}}, \quad
	\fs_0 \sim \initstatedistr[\domparams]{\fs_0}, \\
	\fa_t \sim \pol[\polparams]{\given{\fa_t}{\fs_t}}, \quad
	\domparams \sim \domparamdistr[\domdistrparams]{\domparams},
\end{gathered}
\end{equation}
with the continuous state ${\fs_t \in \stateset[\domparams] \subseteq \RR^{\dimstate}}$ and continuous action ${\fa_t \in \actionset[\domparams] \subseteq \RR^{\dimact}}$ at a time step $t$.
The environment, also called domain, is characterized by its parameters ${\domparams \in \RR^{\dimdomparam}}$ (\eg, masses, friction coefficients, time delays, or camera properties) which are in general assumed to be random variables distributed according to an unknown probability distribution ${\domparamdistrsym \colon \RR^{\dimdomparam} \to \RR^{+}}$.
Here we make the common assumption that the domain parameters or the true system obey a parametric distribution $\domparamdistr[\domdistrparams]{\domparams}$ with unknown parameters $\domdistrparams$ (\eg, mean and variance). 
The domain parameters determine the transition probability density function ${\transprobsym_{\domparams} \colon \stateset[\domparams] \times \actionset[\domparams] \times \stateset[\domparams] \to \RR^{+}}$ that describes the system's stochastic dynamics.
The initial state $\fs_0$ is drawn from the start state distribution ${\initstatedistrsym_{\domparams} \colon \stateset[\domparams] \to \RR^{+}}$.
We model the reward to be a deterministic function of the current state and action ${\rewsym_{\domparams} \colon \stateset[\domparams] \times \actionset[\domparams] \to \RR}$.
Together with the temporal discount factor ${\gamma \in [0,1]}$, the system forms a \ac{MDP} described by the tuple ${\mdp[\domparams] = \tuple{\stateset[\domparams], \actionset[\domparams], \transprobsym_{\domparams}, \initstatedistrsym_{\domparams}, \rewsym_{\domparams}, \gamma}}$.

The goal of a \ac{RL} agent is to maximize the expected (discounted) return, a numeric scoring function which measures the policy's performance.
The expected discounted return of a policy $\pol[\polparams]{\given{\fa_t}{\fs_t}}$ with the parameters $\polparams \in \Theta \subseteq \RR^{\dimpolparam}$ is defined as
\begin{align}
	\label{eq_edr}
	\edr{\polparams,\domparams} =
	&\mathbb{E}_{\fs_0 \sim \initstatedistr*[\domparams]{\fs_0}} \bigg[
	\mathbb{E}_{\fs_{t+1} \sim \transprob*[\domparams]{\fs_t,\fa_t},\ \fa_t \sim \pol[\polparams]{\given{\fa_t}{\fs_t}} } \bigg[ \nonumber\\
	&\sum\nolimits_{t=0}^{T-1} \gamma^t \rewfcn[\domparams]{\fs_t, \fa_t} \Big| \polparams, \domparams, \fs_0 \bigg]
	\bigg].
\end{align}
While learning from experience, the agent adapts its policy parameters.
The resulting state-action-reward tuples are collected in trajectories, \aka rollouts, ${\traj = \set[t=0][T-1]{\fs_t,\fa_t,\rewsym_t} \in \trajspace}$ with $\rewsym_t = \rewfcn[\domparams]{\fs_t, \fa_t}$.
When augmenting the \ac{RL} setting with domain randomization, the goal becomes to maximize the expected (discounted) return for a distribution of domain parameters
\begin{align}
	\label{eq_eedr}
	\eedr{\polparams} &= 
	\Esub{\domparams \sim \domparamdistr*{\domparams}}{\edr{\polparams,\domparams}} \\
	&= \Esub{\domparams \sim \domparamdistr*{\domparams}}{
	   \Esub{\traj \sim p(\traj)}{\sum\nolimits_{t=0}^{T-1} \gamma^t \rewfcn[\domparams]{\fs_t, \fa_t} \Big| \polparams, \domparams, \fs_0}\nonumber%
	}.
\end{align}
The outer expectation with respect to the domain parameter distribution $\domparamdistr{\domparams}$ is the key difference compared to the standard \ac{MDP} formulation.
It enables the learning of robust policies, in the sense that these policies work for a whole set of environments instead of overfitting to a particular instance.

%% file: content/method.tex
The goal of \ac{DiDoR} is to learn a control policy in simulation such that it directly transfers to the real device, i.e., without using any data from the target domain.
To accomplish this, \ac{DiDoR} combines domain randomization with policy distillation.
In the following, we describe the procedure in detail.
A brief summary is given by Algorithm~\ref{algo_pd}.

\begin{figure}[t]
    \centering
    \vspace{0.3em}
    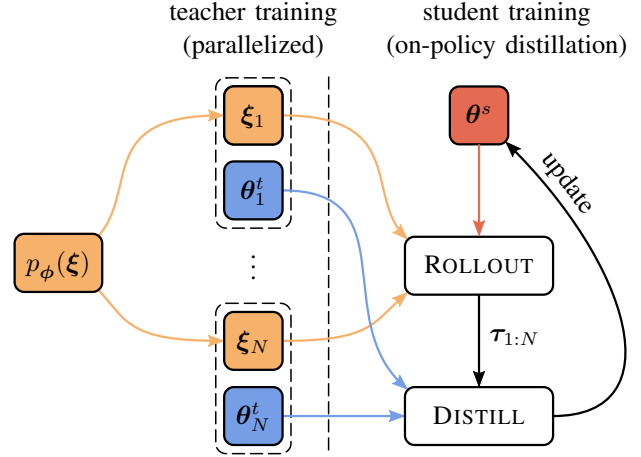
    \caption{Sketch of the proposed sim-to-real method \acs{DiDoR}.}
    \label{fig:DiDoR_sktech}
\end{figure}

\begin{algorithm}[b]
	\caption{\acl{DiDoR}} 
	\label{algo_pd}
	\input{pseudocode/algo_pd.tex}
\end{algorithm}

First, we randomly initialize a set of $N$ teacher policies $\teachpolparams[1:N]$ as well as the student policy $\studpolparams$.
For each teacher, one domain parameter configuration is sampled from the given domain parameter distribution $\domparamdistr[\domdistrparams]{\domparams}$.
The parameters of this distribution are defined a priori by the user.
Next, each teacher is trained exclusively in one of the sampled domains (Figure~\ref{fig:DiDoR_sktech}).
There are no restrictions for the selection of the training algorithm or the policy type.
At this point, we could utilize ensemble methods to create a policy out of the teacher networks, e.g., by averaging their actions. 
However, the time to compute a motor command linearly increases the number of teachers.
Acting in the physical world, robots are subject to hard real-time constraints, which can severely limit ensemble-based methods (Section~\ref{sec:experiments_sim_to_real}).
By distilling all teachers into one student, we obtain a single policy that encodes knowledge from all teachers.
Therefore policy distillation approaches like \ac{DiDoR} have a smaller memory footprint and faster execution times compared to ensemble methods.
To gather the training data for distillation, we execute the student policy in every teacher environment, and collect the rollouts $\ftau_{1:N}$.
Subsequently, we minimize the sum of \ac{KL} divergences between the action distributions of every student-teacher pair, given the collected rollouts:
\begin{equation*}
    \mathcal{L} = \sum\nolimits_{n=1}^{N}
    \mathrm{KL}\!\left( \pol[\teachpolparams[n]]{\ftau_n} \Big\Vert \pol[\studpolparams]{\ftau_n} \right)
\end{equation*}
This loss corresponds to a multi-teacher variant of \emph{on-policy distill}~\cite{Czarnecki_etal_2019} without the entropy penalty.
We found the entropy penalty to be unnecessary in the sim-to-real setting, since the entropy of the resulting student policy is already regulated by the fact that the teacher policies have been learned using domain randomization.
Hence, designing the domain parameter distribution $\domparamdistr[\domdistrparams]{\domparams}$ indirectly controls the entropy of the teacher policies.

%% file: graphics/sketch_DiDoR.pdf_tex
\begingroup%
  \makeatletter%
  \providecommand\color[2][]{%
    \errmessage{(Inkscape) Color is used for the text in Inkscape, but the package 'color.sty' is not loaded}%
    \renewcommand\color[2][]{}%
  }%
  \providecommand\transparent[1]{%
    \errmessage{(Inkscape) Transparency is used (non-zero) for the text in Inkscape, but the package 'transparent.sty' is not loaded}%
    \renewcommand\transparent[1]{}%
  }%
  \providecommand\rotatebox[2]{#2}%
  \newcommand*\fsize{\dimexpr\f@size pt\relax}%
  \newcommand*\lineheight[1]{\fontsize{\fsize}{#1\fsize}\selectfont}%
  \ifx\svgwidth\undefined%
    \setlength{\unitlength}{234.7158309bp}%
    \ifx\svgscale\undefined%
      \relax%
    \else%
      \setlength{\unitlength}{\unitlength * \real{\svgscale}}%
    \fi%
  \else%
    \setlength{\unitlength}{\svgwidth}%
  \fi%
  \global\let\svgwidth\undefined%
  \global\let\svgscale\undefined%
  \makeatother%
  \begin{picture}(1,0.74020545)%
    \lineheight{1}%
    \setlength\tabcolsep{0pt}%
    \put(0,0){\includegraphics[width=\unitlength,page=1]{sketch_DiDoR.pdf}}%
    \put(0.38514473,0.41112385){\color[rgb]{0,0,0}\makebox(0,0)[t]{\lineheight{1.25}\smash{\begin{tabular}[t]{c}$\polparams[1][t]$\end{tabular}}}}%
    \put(0.38514475,0.5318931){\color[rgb]{0,0,0}\makebox(0,0)[t]{\lineheight{1.25}\smash{\begin{tabular}[t]{c}$\domparams_1$\end{tabular}}}}%
    \put(0,0){\includegraphics[width=\unitlength,page=2]{sketch_DiDoR.pdf}}%
    \put(0.38514473,0.04881576){\color[rgb]{0,0,0}\makebox(0,0)[t]{\lineheight{1.25}\smash{\begin{tabular}[t]{c}$\polparams[N][t]$\end{tabular}}}}%
    \put(0.38514474,0.16958509){\color[rgb]{0,0,0}\makebox(0,0)[t]{\lineheight{1.25}\smash{\begin{tabular}[t]{c}$\domparams_N$\end{tabular}}}}%
    \put(0,0){\includegraphics[width=\unitlength,page=3]{sketch_DiDoR.pdf}}%
    \put(0.7474523,0.29170207){\color[rgb]{0,0,0}\makebox(0,0)[t]{\lineheight{1.25}\smash{\begin{tabular}[t]{c}\textsc{Rollout}\end{tabular}}}}%
    \put(0,0){\includegraphics[width=\unitlength,page=4]{sketch_DiDoR.pdf}}%
    \put(0.38514477,0.28180357){\color[rgb]{0,0,0}\makebox(0,0)[t]{\lineheight{1.25}\smash{\begin{tabular}[t]{c}$\vdots$\end{tabular}}}}%
    \put(0,0){\includegraphics[width=\unitlength,page=5]{sketch_DiDoR.pdf}}%
    \put(0.7474523,0.04881611){\color[rgb]{0,0,0}\makebox(0,0)[t]{\lineheight{1.25}\smash{\begin{tabular}[t]{c}\textsc{Distill}\end{tabular}}}}%
    \put(0.74745244,0.53189331){\color[rgb]{0,0,0}\makebox(0,0)[t]{\lineheight{1.25000012}\smash{\begin{tabular}[t]{c}$\polparams[][s]$\end{tabular}}}}%
    \put(0,0){\includegraphics[width=\unitlength,page=6]{sketch_DiDoR.pdf}}%
    \put(0.38514478,0.69739706){\color[rgb]{0,0,0}\makebox(0,0)[t]{\lineheight{1.25}\smash{\begin{tabular}[t]{c}teacher training\\(parallelized)\end{tabular}}}}%
    \put(0.79215284,0.69739706){\color[rgb]{0,0,0}\makebox(0,0)[t]{\lineheight{1.25}\smash{\begin{tabular}[t]{c}student training\\(on-policy distillation)\end{tabular}}}}%
    \put(0,0){\includegraphics[width=\unitlength,page=7]{sketch_DiDoR.pdf}}%
    \put(0.88127037,0.42986273){\color[rgb]{0,0,0}\rotatebox{-52.6545727}{\makebox(0,0)[t]{\lineheight{1.25}\smash{\begin{tabular}[t]{c}update\end{tabular}}}}}%
    \put(0.80493407,0.18236675){\color[rgb]{0,0,0}\makebox(0,0)[t]{\lineheight{1.25}\smash{\begin{tabular}[t]{c}$\trajssym_{1:N}$\end{tabular}}}}%
    \put(0,0){\includegraphics[width=\unitlength,page=8]{sketch_DiDoR.pdf}}%
    \put(0.38514475,0.5318931){\color[rgb]{0,0,0}\makebox(0,0)[t]{\lineheight{1.25}\smash{\begin{tabular}[t]{c}$\domparams_1$\end{tabular}}}}%
    \put(0,0){\includegraphics[width=\unitlength,page=9]{sketch_DiDoR.pdf}}%
    \put(0.07213876,0.29493924){\color[rgb]{0,0,0}\makebox(0,0)[t]{\lineheight{1.25}\smash{\begin{tabular}[t]{c}$\domparamdistr[\domdistrparams]{\domparams}$\end{tabular}}}}%
  \end{picture}%
\endgroup%

%% file: pseudocode/algo_pd.tex
\DontPrintSemicolon
\Input{%
domain parameter distribution $\domparamdistr[\domdistrparams]{\domparams}$,
number of teachers $N$,
iterations $I$, and
epochs $E$
}
\Output{%
trained student policy parameters $\studpolparams$
}
Initialize the teacher $\teachpolparams[1:N]$ and the student $\studpolparams$ policy parameters randomly\;
Sample one environment instance per teacher $\domparams_{1:N} \sim \domparamdistr[\domdistrparams]{\domparams}$\;
Train the teachers individually on their environment
$\teachpolparams[1:N] \gets \textsc{PolOpt}(\domparams_{1:N})$ \Comment*[r]{parallelizable}
\For{each iteration $i = 1\!:\!I$}{
    \For(\Comment*[f]{parallelizable}){each teacher $n = 1\!:\!N$}{
        Execute student in each teacher domain
        $\ftau_n \gets \textsc{Rollout}(\polsym_{\studpolparams}, \domparams_n)$\Comment*[f]{data collection}
    }
    \For(\Comment*[f]{policy distillation}){each epoch $e = 1\!:\!E$} {
        Compute the policy distillation loss
        $\mathcal{L} \gets \sum_{n=1}^{N} \KL*{\pol[\teachpolparams[n]]{\ftau_n}} {\pol[\studpolparams]{\ftau_n}}$\;
        Update $\studpolparams$ using $\KLlosssym$\;
    }
}

%% file: content/experiments.tex
To examine the robustness of \ac{DiDoR}, we conducted three \simtosim (Section~\ref{sec:experiments_sim_to_sim}) as well as two \simtoreal experiments (Section~\ref{sec:experiments_sim_to_real}).
The \simtosim scenarios can be seen as a proof of concept, where we evaluate the learned policies in 32 unknown simulation instances randomly sampled from the domain parameter distribution.
The \simtoreal experiments are significantly more challenging due to the underactuated nature of the chosen tasks as well as the inevitable approximation errors of the simulator.
Moreover, robot control imposes hard time constraints.
Since the control cycles of the cart-pole and the Furuta pendulum (Figure~\ref{fig:platforms}) are running at \SI{500}{\hertz}, the policy must not take more that \SI{2}{\milli\second} to compute one forward pass.
For simulating the randomized dynamical systems, we use the MuJoCo physics engine~\cite{Todorov_etal_2012} as well as the SimuRLacra framework~\cite{Muratore_SimuRLacra}. 

We compare \ac{DiDoR} to the following baseline methods:
\begin{enumerate}
    \item \textbf{\acf{UDR}} randomly samples a new set of domain parameters at every iteration, and updates the policy based on the data generated using these parameters.
    Thus, the policy permanently experiences new instances of the domain, which in turn leads to an increased robustness against model mismatch.
    \item \textbf{Ensemble Methods} combine a set of policies in order to obtain a stronger model.
    Since we trained several teachers for DiDoR on different (random) domain instances, we can use them for the formation of an ensemble. However,  there are numerous ways to query them.
    Motivated by prior experiments, we chose to average the action outputs of all policies within the ensemble.
    Hence, this baseline serves as a straightforward ablation of \ac{DiDoR} without the policy distillation step.
    A considerable disadvantage of ensemble methods is that the execution time scales linearly with the number of elements, since this method requires a forward pass through all teacher policies. Nevertheless, this effect can be mitigated by using parallelization.
    \item \textbf{\acf{P2PDRL}}~\cite{Zhao2020} is a policy distillation method aiming at learning robust policies.
    Similar to \ac{DiDoR}, several teachers, here called workers, are trained on different domains.
    However, while our teachers are trained independently, these workers are trained at the same time and share knowledge via a special distillation loss, which in the end aligns the action distributions of the policies.
    Moreover, the domain parameters of the environments of the workers are drawn each iteration, while in DiDoR they stay fixed.
\end{enumerate}
The training method as well as the policy architecture can be chosen freely for all algorithms.
We decided to use \ac{PPO}~\cite{Schulman_etal_2017} and feedforward neural networks.
To facilitate a fair comparison, we tried to train all methods with similar configurations, e.g., \ac{UDR} got as many iterations as all teachers combined.
However, there were practical limitations which made a fair comparison difficult.
While the teachers in \ac{DiDoR} can train independently, all workers in \ac{P2PDRL} work simultaneously and need to be kept in memory at the same time.
Therefore, we had to reduce the number of workers for \ac{P2PDRL}, especially for the MuJoCo environments (Section~\ref{sec:experiments_sim_to_sim}) to one fourth of the number of teachers in \ac{DiDoR}. 
Moreover, we choose all hyper-parameters which are common for the selected approaches, e.g. the policy architecture, identically.
A report all hyper-parameters for training can be found in Table~\ref{tab:training_param}.
Moreover, we list the domain parameter distributions from the \simtoreal experiments in the appendix.
To accelerate the learning, we followed the recommendations from~\cite{Andrychowicz_2020} for on-policy training, finding that many of the suggestions did not yield the desired effect.
However, initializing the action distribution of the policy such that it had approximately zero mean, significantly increased the initial return, thus decreased the number of necessary iterations.

\subsection{Sim-to-Sim Experiments}
\label{sec:experiments_sim_to_sim}
\begin{figure*}[t]
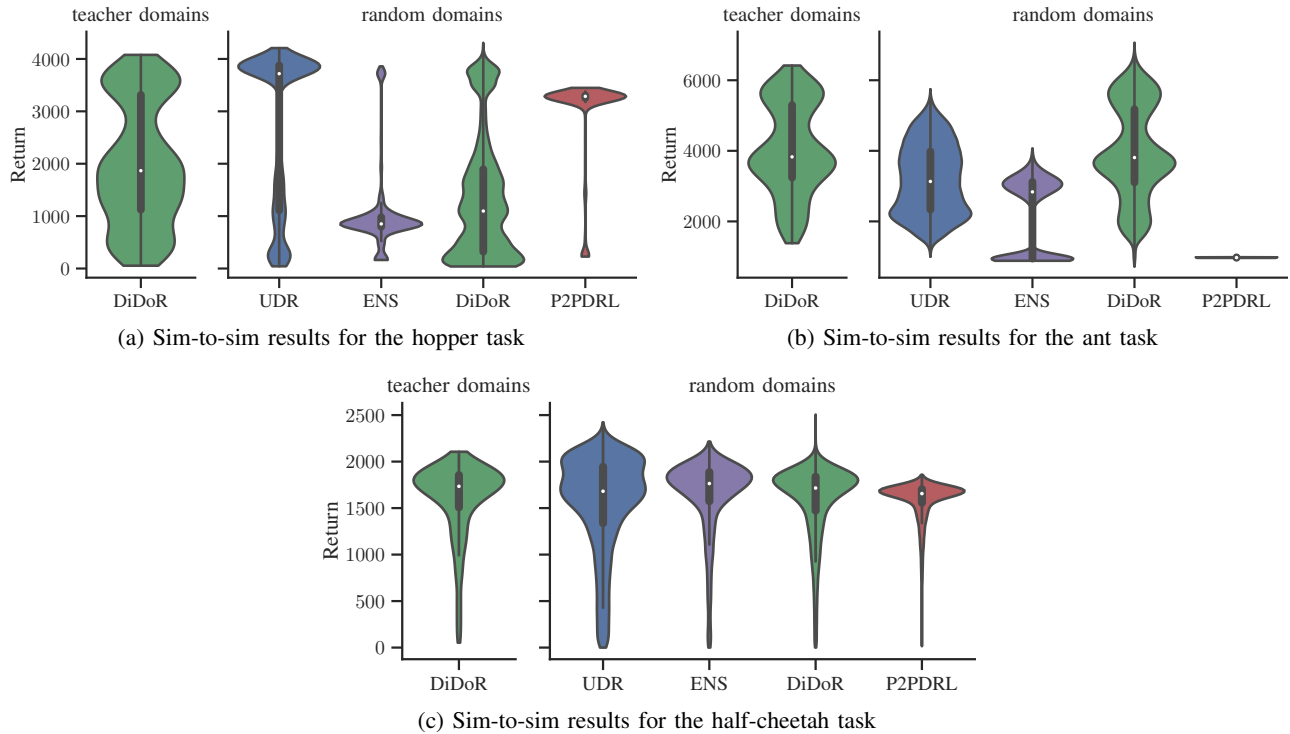

    \centering
    \begin{subfigure}{0.48\linewidth}
        \centering
        \resizebox {\linewidth} {!} {
            \input{graphics/hop_oldP2P_rel_1000_plot.pgf}
        }
        \caption{\Simtosim results for the hopper task\vspace{1.0em}}
    \end{subfigure}
    \begin{subfigure}{0.48\linewidth}
        \centering
        \resizebox {\linewidth} {!} {
            \input{graphics/ant_oldP2P_rel_1000_plot.pgf}
        }
        \caption{\Simtosim results for the ant task\vspace{1.0em}}
    \end{subfigure}
    \begin{subfigure}{\textwidth}
        \centering
        \resizebox {0.5\linewidth} {!} {
            \input{graphics/cth_oldP2P_rel_1000_plot.pgf}
        }
        \caption{\Simtosim results for the half-cheetah task}
    \end{subfigure}
    \caption{%
    \Simtosim performance of the proposed method (\acs{DiDoR}) as well as the baselines in three simulation environments:
    (left) evaluation of \acs{DiDoR}'s student policy in its teacher domains.
    (right) evaluation of all methods in randomly sampled domains which have not been seen during training.
    The results were obtained by evaluating four policies, trained from different random seeds, per algorithm, each executing 100 rollouts with different initial states on 32 unknown simulation instances.
    The white circles mark the median values.
    }
    \label{fig:sim_to_sim_experiment_result}
\end{figure*}

For a first validation in simulation, we used three locomotion tasks from the OpenAI Gym~\cite{openai_gym}: hopper, half-cheetah, and ant.
We augmented each MuJoCo environment with domain randomization.
However, the domain parameters of the training domain have lower variance than those of the test domain, therefore we can evaluate the robustness of the transfer. 
%
The commonality of the rather mixed \simtosim results displayed in Figure~\ref{fig:sim_to_sim_experiment_result} is that \ac{DiDoR} performs en par or better than the baselines depending on which metric is applied (e.g., highest median or highest peak performance).
From the fact that, except for \ac{UDR}, every approach works on at least two out of the three tasks, we conclude that the high variance of the returns originates from too aggressive domain randomization, i.e., a too broad domain parameter distribution.


\subsection{Sim-to-Real Experiments}
\label{sec:experiments_sim_to_real}
\begin{figure*}[t]
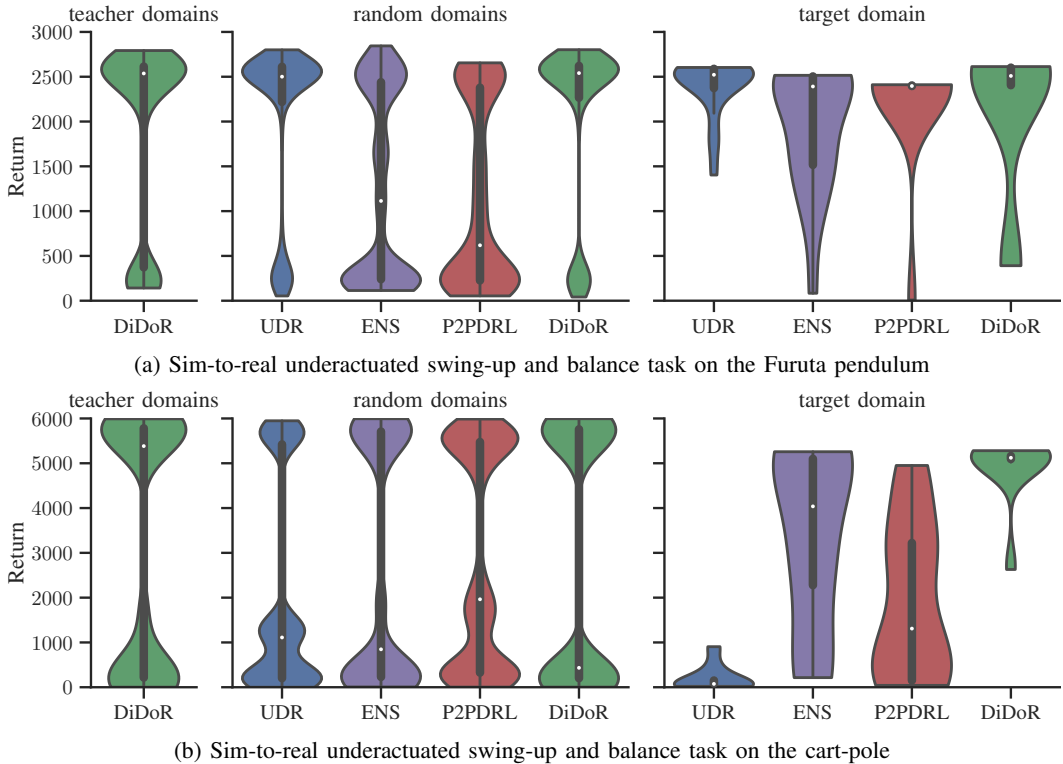

    \centering
    \begin{subfigure}{\textwidth}
        \centering
        \resizebox {0.8\linewidth} {!} {
            \input{graphics/qq-su_3000_plot.pgf}
        }
        \caption{\Simtoreal underactuated swing-up and balance task on the Furuta pendulum\vspace{0.5em}}
        \label{fig:eval_qq}
    \end{subfigure}
    \vfil
    \begin{subfigure}{\textwidth}
        \centering
        \resizebox {0.8\linewidth} {!} {
            \input{graphics/qcp-su_6000_plot.pgf}
        }
        \caption{\Simtoreal underactuated swing-up and balance task on the cart-pole}
        \label{fig:eval_qcp}
    \end{subfigure}
    \caption{%
    \Simtoreal performance of the proposed method (\acs{DiDoR}) as well as the baselines on two  physical platforms:
    (left) evaluation of \acs{DiDoR}'s student policy in its teacher domains.
    (middle) evaluation of all methods in randomly sampled domains which have not been seen during training.
    The results were obtained by evaluating four policies, trained from different random seeds, per algorithm, each executing 100 rollouts with different initial states on 32 simulation instances.
    The white circles mark the median values.
    (right) evaluation of all methods on the real-world device.
    The return was estimated from four policies per method, each executed 10 times on both platforms.
    }
    \label{fig:sim_to_real_experiment_result}
\end{figure*}

We evaluate the chosen methods' zero-shot \simtoreal transfer on two underactuated swing-up and balance tasks (Figure~\ref{fig:platforms}).
The Furuta pendulum~\cite{Furuta_Kobayashi_92} is a rotary inverted pendulum with fast dynamics.
Its relatively short pendulum arm makes the stabilization of the upper equilibrium difficult.
The cart-pole is a cart-driven inverted pendulum on a rail.
For this platform, the contact between the cart's wheel (soft plastic) and the rail (metal) is particularly hard to model.
We strongly believe that the simulation does not capture some of the highly nonlinear friction effects.
This hypotheses is backed up by the results in Figure~\ref{fig:eval_qcp} (right), which show a higher variance, i.e., a higher failure rate, across all methods.
In contrast, the middle subplots in Figure~\ref{fig:sim_to_real_experiment_result} reveal that all methods were able to solve both simulated swing-up and balance tasks.
Comparing the left and middle subplots in Figure~\ref{fig:sim_to_real_experiment_result}, we observe that \ac{DiDoR} performed equally well on randomly sampled domains as on the teacher domains, providing evidence for good generalization capabilities.
We explain the \ac{UDR} policies' inability to transfer by a too narrow domain parameter distribution.
The most likely consequence of this is that \ac{UDR} did not capture enough variation during training while still learning with its high-variance gradient estimates.
The \simtoreal deployment on the Furuta pendulum is more benign (Figure~\ref{fig:eval_qq}).
Aside from negligibly few failures of two baseline methods, the vast majority of policies transferred.

During our experiments, we observed that at some points in time the individual elements of the ensemble baselines commanded actions which approximately canceled each other.
If desired, this situation can be resolved by replacing the average over policy outputs with a different querying strategy. 
Finally, we attribute the relatively low performance of \ac{P2PDRL} to a suboptimal hyper-parameter selection.
Unfortunately, the suggestions in~\cite{Zhao2020} did not produce satisfying results.
This might be due to the fact that the authors tuned the parameters for other \simtosim tasks.

\begin{table}[bth]
    \centering
    \caption{Hyper-parameters for training all reported policies.}
    \label{tab:training_param}
    
    \begin{tabular}{ll}
      \toprule
      Parameter & Value \\
      \midrule
        algorithm & PPO~\cite{Schulman_etal_2017} with GAE~\cite{Schulman_etal_2015}\\
        policy architecture & FNN 64-64 tanh\\
        rollout length & 8000 steps at \SI{500}{\hertz}\\
        discount factor $\gamma$ & 0.99\\
        lambda & 0.97\\
        parallel environments & 48 (\simtoreal), 64 (\simtosim)\\
        number of iterations & 40\\
        initial exploration variance & 1.0\\
        PPO clip ratio & 0.1\\
        learning rate & \num{1e-3} (\simtoreal), \num{1e-4} (\simtosim)\\
        KL clip (PPO update) & 0.05\\
      \midrule
      \multicolumn{2}{c}{\ac{DiDoR} specific}\\
      teacher number & 16 (\simtoreal), 8 (\simtosim)\\
      \midrule
      \multicolumn{2}{c}{\ac{P2PDRL} specific}\\
      worker number & 4 (cart-pole), 2 (all other tasks)\\
      distillation loss coefficient $\alpha$ & 0.05 (\simtoreal), 0.01 (\simtosim)\\
      \bottomrule
    \end{tabular}
\end{table}

%% file: content/conclusion.tex
We have introduced \acf{DiDoR}, an algorithm that combines domain randomization and knowledge distillation to learn policies that are able to transfer from simulation to reality.
By distilling the expert knowledge of multiple teachers, which have been learned in various instances of the problem, the resulting student policy is more robust to the inevitable model mismatch that occurs when deploying a learned policy on the physical system.
We have evaluated the presented method on three sim-to-sim and two sim-to-real experiments.
Our results show the advantages of \ac{DiDoR} over uniform domain randomization, an ensemble of teachers, and \acl{P2PDRL}.
In contrast to these baselines, \acs{DiDoR} neither increases the required memory capacity, nor the time to compute an action.
Both features make the proposed approach better suited for high-frequency control on real robots.
Moreover, \acs{DiDoR} is vastly parallelizable and has been shown to generalize better to unseen domains.
For future research, we plan to investigate the effect of replacing the distillation loss, as for example described in~\cite{Czarnecki_etal_2019}.

%% file: content/appendix.tex
In the subsequent tables, we list the domain parameter distributions of the Furuta pendulum (Table~\ref{tab:domain_param_qq}) and the cart-pole (Table~\ref{tab:domain_param_qcp}) experiments.
Normal distributions are parameterized with mean and standard deviation, uniform distributions with lower and upper bound.

\begin{table}
    \centering
    \caption{The domain parameter distribution for the \simtoreal Furuta pendulum experiment.}
    \label{tab:domain_param_qq}
    \begin{tabular}{ll}
    	\toprule
    	Parameter                 & Distribution \\
    	\midrule
    	gravity constant          & $\distrnormal{g}{\num{9.81},\num{0.981}} \si{\kilogram}$ \\
    	pend. pole mass           & $\distrnormal{m_p}{\num{0.024},\num{0.048}} \si{\kilogram}$ \\
    	rot. pole mass            & $\distrnormal{m_p}{\num{0.095},\num{0.019}} \si{\kilogram}$ \\
        pend. pole length         & $\distrnormal{l_p}{\num{0.129},\num{0.026}} \si{\meter}$ \\
    	rot. pole length          & $\distrnormal{l_r}{\num{0.085},\num{0.017}} \si{\meter}$ \\
        pend. pole damping        & $\distruniform{d_p}{\num{1e-6},\num{2.5e-7}} \si{\newton\meter\second\per\radian}$ \\
    	rot. pole damping         & $\distrnormal{d_r}{\num{5e-6},\num{1.25e-6}} \si{\newton\meter\second\per\radian}$ \\
    	motor resistance          & $\distrnormal{R_m}{\num{8.4},\num{1.68}} \si{\ohm}$ \\
    	motor constant            & $\distrnormal{k_m}{\num{0.042},\num{8.4e-3}} \si{\newton\meter\per\ampere}$ \\
        \bottomrule
    \end{tabular}
\end{table}
    
\begin{table}
    \centering
    \caption{The domain parameter distribution for the \simtoreal cart-pole experiment.}
    \label{tab:domain_param_qcp}
    \begin{tabular}{ll}
    	\toprule
    	Parameter                 & Distribution \\
    	\midrule
    	gravity constant          & $\distrnormal{g}{\num{9.81},\num{0.981}} \si{\kilogram}$ \\
    	cart mass                 & $\distrnormal{m_c}{\num{0.38},\num{0.076}} \si{\kilogram}$ \\
    	pole mass                 & $\distrnormal{m_p}{\num{0.127},\num{2.54e-2}} \si{\kilogram}$ \\
        pole length               & $\distrnormal{l_p}{\num{0.16825},\num{3.365e-2}} \si{\meter}$ \\
    	rail length               & $\distrnormal{l_r}{\num{0.814},\num{0.163}} \si{\meter}$ \\
    	motor pinion radius       & $\distrnormal{r_{mp}}{\num{6.35e-3},\num{1.27e-3}} \si{\meter}$ \\
    	gear ratio                & $\distrnormal{K_g}{3.71,0.93}$ \\
    	gearbox efficiency        & $\distruniform{\eta_g}{\num{0.675},\num{1.0}}$ \\
    	motor efficiency          & $\distruniform{\eta_m}{\num{0.675},\num{1.0}}$ \\
    	motor moment of inertia   & $\distrnormal{J_m}{\num{3.9e-7},\num{9.75e-8}} \si{\kilogram\meter\squared}$ \\
    	motor torque constant     & $\distrnormal{k_m}{\num{7.67e-3},\num{1.92e-3}} \si{\newton\meter\per\ampere}$ \\
    	motor armature resistance & $\distrnormal{R_m}{\num{2.6},\num{0.65}} \si{\ohm}$ \\
    	\makecell[l]{motor viscous damping \\  coeff. \wrt load} & $\distruniform{B_{\textrm{eq}}}{\num{4.05},\num{6.75}} \si{\newton\second\per\meter}$ \\
    	pole viscous friction coeff.           & $\distruniform{B_p}{\num{1.8e-3},\num{3.0}} \si{\newton\second}$ \\
    	friction coefficient cart - rail & $\distruniform{\mu_c}{\num{0.01},\num{0.03}}$ \\
        \bottomrule
    \end{tabular}
\end{table}